\documentclass[letterpaper, 10 pt, conference]{ieeeconf}  

\usepackage{color}
\newcommand{\ie}{i.e.\ }
\newcommand{\eg}{e.g.\ }
\newcommand{\Reffig}[1]{Figure~\ref{#1}}

\newcommand{\Refeq}[1]{Equation~\ref{#1}}
\newcommand{\Reftab}[1]{Table~\ref{#1}}

\usepackage{graphics} 
\usepackage{epsfig} 
\usepackage{algorithm}
\usepackage{algpseudocode}
\usepackage{amsmath}

\usepackage{booktabs}
\usepackage{multirow}
\usepackage{multicol}
\usepackage{makecell}
\usepackage{footnote}

\usepackage{graphicx}
\usepackage{caption2}
\usepackage{subfigure}
\usepackage{float}
\usepackage{cite}
\usepackage[colorlinks,linkcolor=blue]{hyperref}

\title{\LARGE \bf
Occlusion Aware Unsupervised Learning of Optical Flow From Video
}

\author{Jianfeng Li$^{3}$, Junqiao Zhao$^{*, 1, 2, 4}$, Tiantian Feng$^{3}$, Chen Ye$^{1, 2}$, Lu Xiong$^{4}$ 
\thanks{*This work is supported by the National Key Research and Development Program of China (No. 2018YFB0105103, No. 2017YFA0603104), the National Natural Science Foundation of China (No. U1764261, No. 41801335, No. 41871370), the Natural Science Foundation of Shanghai (No. kz170020173571, No. 16DZ1100701) and the Fundamental Research Funds for the Central Universities (No. 22120180095).}
\thanks{$^{1}$The Key Laboratory of Embedded System and Service Computing, Ministry of Education, Tongji University, Shanghai
		{\tt\small zhaojunqiao@tongji.edu.cn}}%
\thanks{$^{2}$Department of Computer Science and Technology, School of Electronics and Information Engineering, Tongji University, Shanghai}%
\thanks{$^{3}$School of Surveying and Geoinfomatics, Tongji University, Shanghai}
\thanks{$^{4}$Insititute of Intelligent Vehicle, Tongji University, Shanghai}
}

\begin{document}
\maketitle
\thispagestyle{empty}
\pagestyle{empty}

\begin{abstract}
    In this paper, we proposed an unsupervised learning method for estimating the optical flow between video frames, especially to solve the occlusion problem.
    Occlusion is caused by the movement of an object or the movement of the camera, defined as when certain pixels are visible in one video frame but not in adjacent frames.
    Due to the lack of pixel correspondence between frames in the occluded area, incorrect photometric loss calculation can mislead the optical flow training process.
    In the video sequence, we found that the occlusion in the forward ($t\rightarrow t+1$) and backward ($t\rightarrow t-1$) frame pairs are usually complementary.
    That is, pixels that are occluded in subsequent frames are often not occluded in the previous frame and vice versa.
    Therefore, by using this complementarity, a new weighted loss is proposed to solve the occlusion problem.
    In addition, we calculate gradients in multiple directions to provide richer supervision information.
    Our method achieves competitive optical flow accuracy compared to the baseline and some supervised methods on KITTI 2012 and 2015 benchmarks.
    This source code has been released at \url{https://github.com/jianfenglihg/UnOpticalFlow.git}.
\end{abstract}

\section{INTRODUCTION}
Optical flow estimation is a fundamental problem in computer vision \cite{AubertComputing}.
Traditional methods include region-based matching, energy-based methods, and phase-based techniques, etc. \cite{Chao2013}.
\cite{Fischer} first introduced a convolutional neural network (CNN)-based method into the optical flow estimation.
The results show this end-to-end method can obtain dense results with high framerate.
Nevertheless, most methods train the neural network in a supervised manner.
It is known that the ground truth of optical flow is hard to obtain for real video sequences.
And there are few sensors that can obtain optical flow directly \cite{Janai2017}.
As a result, many methods use synthetic data for training \cite{Mayer2016A,Fischer}, which limits the generalization ability of these supervised approaches.

To overcome the limitation, unsupervised methods are proposed \cite{Ren2017}.
The unsupervised learning optimizes a photometric loss between a target frame and the warped adjacent frame with the predicted optical flow, as an alternative to directly optimize the loss between the predicted optical flow and the ground truth. 
With this unsupervised learning method, almost any video sequence can be employed as a training dataset.
However, an important factor affecting the accuracy of the unsupervised method is occlusion.

Occlusion is defined as when certain pixels are visible in one video frame but not in adjacent frames.
The photometric loss in the temporarily occluded region is undefined because there is no pixel correspondence between frames.
Incorporating the photometric loss belonging to the occluded area can interfere with or even mislead the training of the neural network.
Therefore, how to handle the occlusion problem is explored in many recent studies \cite{Janai2018,Meister2017,Wang}.

There are two ways to deal with this problem. 
One is the estimation of an occlusion mask jointly with the prediction of the optical flow using the neural network \cite{Janai2018}.
The other is the generation of the occlusion mask based on prior knowledge \cite{Wang,Meister2017,UnFLOWICIP}, \eg{forward-backward consistency} \cite{Sundaram2010Dense}.
While both ways improve the performance of unsupervised learning, the latter is more reliable with the well-established prior knowledge.
However, existing methods often require fine-tuning the occlusion threshold for a specific dataset, which limits their ability to generalize in a wide range of datasets.

In this paper, we introduce a new prior knowledge that the occlusion is complementary in the forward ($t\rightarrow t+1$) and backward ($t\rightarrow t-1$) frame pairs.
Specifically, with the assumption that the camera frame rate is sufficiently high and the motion is continuous, the pixels on the intermediate frame can always find corresponding pixels in either the previous or the subsequent frame (being occluded), or in both (not occluded).
That is, pixels that are occluded in one direction, \eg{forward direction}, of the video are not occluded in the opposite direction, \eg{backward direction}.
Therefore, the photometric loss calculated in the video direction where occlusion occurs will be greater than that in the opposite direction.

If only the smaller one between the forward loss and the backward loss is retained for each pixel, the occlusion will be excluded from the total loss, as shown in \cite{Godard2018}.
However, this also discards meaningful supervision information from non-occluded areas.
To solve this problem, we proposed a soft weighting method to weaken the influence of the occluded pixels on loss calculation.
In addition, we use a second-order photometric loss to handle light changes, and we calculate the gradient in multiple directions, so the supervised information will be richer especially at the edges.
We validated the proposed method in the KITTI benchmark \cite{Geiger2012Are}, which achieves competitive optical flow accuracy compared to the baseline.

We summarize the contributions of this paper as follows:
\begin{itemize}
    \item A novel pixel-wise weighting method is proposed to handle occlusion using a new prior knowledge that occlusion is complementary in forward and backward frame pairs.
    \item We added multiple gradient directions when calculating second-order photometric loss, which improves the optical flow estimation at the edges. 
    \item The proposed method excludes the tuning of occlusion related thresholds
\end{itemize}

\section{RELATED WORKS}

\subsection{Supervised Learning of Optical Flow}
Classic methods for calculating optical flow are based on the assumptions of small pixel displacements and brightness consistency, but there is no such strict restriction for deep supervised learning methods.
Flownet \cite{Fischer} first uses CNN to predict dense optical flow with an encoder-decoder architecture.
And it proposes two encoder structures, \ie{FlownetS and FlownetC}.
The FlownetS stacks two input images directly and the FlownetC adds a layer that correlates feature vectors at different image locations.
However, the performance of Flownet on KITTI benchmark is worse than that on synthetic MPI Sintel benchmark \cite{Butler:ECCV:2012}.
Inspired by the iterative refinement, Flownet2 \cite{Ilg} proposes to stack multiple networks for optical flow estimation.
Compared to Flownet, Flownet2 decreases the estimation error by more than 50\%.

Spynet \cite{RanjanOptical} constructs the image pyramid and estimates optical flow in a coarse-to-fine manner.
At each level, the estimated optical flow from the upper coarser level is used to warp one of the scaled input images of this level.
And the warped image and the reference image are fed into the network to estimate the residual optical flow. 
The output optical flow of this level is obtained by adding the residual flow with the optical flow from the upper level. 
Spynet is 96\% smaller than FlowNet in terms of model parameters, but is less accurate than FlowNet2.

Pwcnet\cite{SunPWC} also adopts the pyramid strategy.
Unlike Spynet, Pwcnet uses feature spatial pyramid and feature warping instead.
In addition, the output of Pwcnet is the optical flow at each level.
Pwcnet both increased the accuracy and reduced the size of parameters, which is about 17 times smaller than FlowNet2.

LiteFlowNet\cite{Hui} shares a similar architecture with the Pwcnet, which is built up by the coarse-to-fine architecture, feature warping, and cost volume for optical flow estimation.
LiteFlowNet estimates the residual flow at each pyramid level thus the model size is smaller than Pwcnet.
However, both methods perform equally in KITTI and Sintel benchmarks. 

\subsection{Unsupervised Learning of Optical Flow}

Recently, the unsupervised methods have been intensively studied because they do not rely on labeled data.
The key to unsupervised learning is the use of image reconstruction loss, which penalties the difference between the target image and the warped target image calculated by the estimated optical flow and source images.
\cite{PatrauceanSpatio, Ren2017, Yu2016} first use photometric and smooth loss to achieve unsupervised learning for optical flow estimation. 
In addition to basic photometric and smooth loss, \cite{Zhong2019} introduces the epipolar constraint from the two-view geometry to the unsupervised framework.
They propose a low-rank constraint which provides a way of regularizing the optical flow estimation.
\cite{Zhu2019} adds a structural similarity (SSIM) \cite{Wang2004Image} loss in addition to the photometric and smooth loss.
SSIM helps the network to learn the structural information between frames.
This provides extra supervision information to train the neural network, but SSIM requires intensive computation.

Another interesting route of unsupervised learning is the combination of the optical flow estimation with depth and ego-motion estimation.
In classic multi-view geometry, the three are closely correlated with each other.
\cite{Yin2018,Ranjan2018,Zhang2019,Wang2019,Teng2018,Zou2018,Chen2019,Wang2018,Mayer2016A} unify the optical flow, depth and ego-motion estimation under the unsupervised framework, making them mutually constrained.

\subsubsection*{Occlusion}
As suggested by many recent studies \cite{Baker2011}, handling occlusion is a major factor that can improve the unsupervised learning performance apart from adding more constraints.
Back2Future \cite{Janai2018} leverages three frames to estimate the forward and the backward optical flow and the occlusion mask respectively.
UnFlow \cite{Meister2017} leverages two frames to estimate the forward ($t\rightarrow t+1$) and the backward optical ($t+1\rightarrow t$) flow, but detecting occlusion according to the forward-backward consistency prior which is that the forward flow should be the inverse of backward flow in non-occluded areas.
When the mismatch between these two flows is too large and over a threshold, UnFlow mark pixels as occluded.
OAFlow \cite{Wang} also feds two frames into the network and estimate the forward and backward flow by sharing weights.
They create a range map using forward warping and the occlusion map can be obtained by simply thresholding the range map.
The occlusion map predicted by the neural network is noisier than that calculated based on prior knowledge.
However, most methods that utilize prior knowledge need a fine-tune occlusion threshold for a specific dataset, making it difficult to train simultaneously on data from multiple sources. 

\section{Method}

\subsection{Overall Structure}
\begin{figure*}[tbp]
    \centering
    \includegraphics[width=0.8\textwidth]{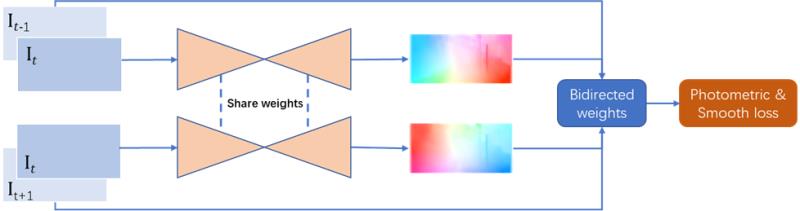}
    \caption{The overall structure of the proposed system}
    \label{overall}   
\end{figure*}

The overall structure is shown in \Reffig{overall}.
The structure of our neural network is similar to Pwcnet \cite{SunPWC} which is lightweight and relies on only two input images.
During training phase, given three adjacent frames \{$I_{t-1}$,$I_{t}$,$I_{t+1}$\}, we feed \{$I_{t}$,$I_{t-1}$\} and \{$I_{t}$,$I_{t+1}$\} into the network respectively by sharing weights, and then the bidirectional optical flow, \ie{the backward flow $F_{b}$ and the forward flow $F_{f}$} are obtained.
For each direction, we use the estimated optical flow and a source image, either $I_{t-1}$ or $I_{t+1}$, to reconstruct a warped target image $I'_{t}$.
Then the photometric error is calculated between $I_{t}$ and $I'_{t}$ and the smooth regularization is employed.

The total loss contains the bidirected photometric loss $L_{p}$ and the bidirected smooth loss $L_{s}$.
Moreover, we utilize the difference of photometric error between the forward and backward directions to calculate occlusion maps.
The bidirected losses are then weighted according to the occlusion map.

\subsection{Photometric Loss and Smooth Regularization}

\subsubsection*{Photometric Loss}
The parameters of the network are learned by jointly minimizing the loss:
\begin{equation}
    \label{E1}
    L = \lambda_{p}L_{p} + \lambda_{s}L_{s}.
\end{equation}
where ${\lambda_{p},\lambda_{s}}$ are the weights on the respective loss term.
$L_{p}$ and $L_{s}$ represent photometric loss and smooth regularization.
Similar to \cite{Ren2017,Brox2004High}, we not only use the first order photometric error, but also the second-order error, because the latter is more robust to light change than the former, especially in natural scenes like in KITTI \cite{Geiger2012Are}.

The first order photometric loss is defined as follows:
\begin{equation}
    \begin{aligned}
        E_{b}(p)&= I_{t}(p)-I_{t-1}(F_{b}(p)+p)\\
        E_{f}(p)&= I_{t}(p)-I_{t+1}(F_{f}(p)+p)
    \end{aligned}
\end{equation}

\begin{equation}
    \label{E2}
    \begin{aligned}
    L_{p1st}&= \sum_{p}\{\delta(E_{b}(p))+\delta(E_{f}(p))\}
    \end{aligned}
\end{equation}
Where $E_{b}$ is backward photometric error, and $E_{f}$ is forward photometric error.
And $p$ represents the image coordinate, and $\delta(.)$ is the robust Charbonnier function: $\delta(x)=(x^{2}+\epsilon^{2})^{\kappa}$.

%

The second-order photometric loss is defined as the following:
\begin{equation}
    \label{E3}
    \begin{aligned}
    L_{p2nd}&= \sum_{d}\sum_{p}\{\delta(\nabla ^{d}I_{t}(p)-\nabla ^{d}I_{t-1}(F_{b}(p)+p))\\
            &+\delta(\nabla ^{d}I_{t}(p)-\nabla ^{d}I_{t+1}(F_{f}(p)+p))\}
    \end{aligned}
\end{equation}
where $\nabla^d$ means calculate gradient of image along the direction $d$, taking the x direction as an example, $\nabla ^{x}I(i,j)=I(i,j)-I(i-1,j)$.
We calculate the gradient loss in four directions, with the angles of 0, 45, 90, 180 degrees.
Finally, our photometric loss is:
\begin{equation}
    \label{E4}
    L_{p}=\lambda_{p1st}L_{p1st} + \lambda_{p2nd}L_{p2nd}
\end{equation}

\begin{figure}[tp]
    \centering
    \includegraphics[width=0.45\textwidth]{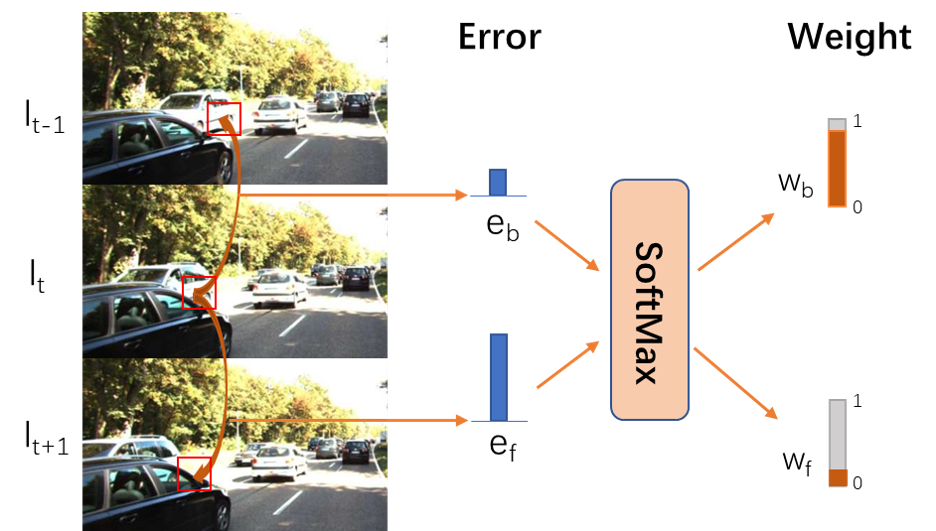}
    \caption{The illustration of our weighting method.}
    \label{min}   
\end{figure}

\subsubsection*{Smooth Regularization}
The way of smooth regularization is similar to \cite{Wang}, we also use second-order smooth regularization.
And we adopt edge-aware formulation similar to \cite{Janai2018}.
Our smoothed loss function is defined as follows:
\begin{equation}
    \begin{aligned}
        \label{E5}
        L_{s2nd}&= \sum_{d}\sum_{p}(\vert\nabla ^{d}F_{b}(p)\vert^{2} e^{-\alpha\vert\nabla ^{d}I_{t}(p)\vert}\\
            &+\vert\nabla ^{d}F_{f}(p)\vert^{2} e^{-\alpha\vert\nabla ^{d}I_{t}(p)\vert})
    \end{aligned}
\end{equation}
where $\alpha$ controls the weight of smooth on the edge.

\subsection{Weighted Occlusion}

\label{section:occ}
\begin{figure}[htbp]
    \centering
    \includegraphics[width=0.3\textwidth]{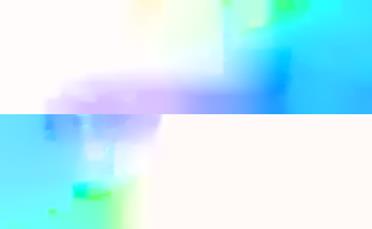}
    \caption{The demonstration of the artefacts when using the minimum photometric error between the forward and backward directions. Only half the position of the image is learned. Up: the backward optical flow. Below: the forward optical flow.}
    \label{min_example}   
\end{figure}

Occluded pixels can not provide valid supervision information.
An intuitive idea is to detect the occluded pixels and to remove them.
As illustrated in \Reffig{min}, the pixel in image $I_{t}$ will be occluded in image $I_{t+1}$, but can be found in image $I_{t-1}$.
Therefore, the photometric error of the pixels between $I_{t}$ and $I_{t+1}$ is greater than that between $I_{t}$ and $I_{t-1}$.
According to this fact, we propose a method to reduce the weight of photometric loss belonging to the occluded area, and improve the weight of photometric loss belonging to the non-occluded area.
\cite{Godard2018} also utilizes this fact to deal with occlusion while predicting the depth from a video.
But they just throw away the bigger error in photometric loss between forward and backward directions, \ie{$min\{E_b(p),E_f(p)\}$}.
However, applying that method on optical flow estimation will cause another problem.
Dropping larger errors directly result in the optical flow in the half position of the image cannot be learned, as demonstrated in \Reffig{min_example}.
The reason is that in non-occluded areas, the supervised information provided by the two directions is redundant for predicting the depth, but is necessary for predicting the optical flow.
Therefore, we propose a soft way to deal with occlusion.
We weight the forward and backward photometric loss according to their relative magnitude.
The weight can be calculated using the Softmax function:
\begin{equation}
    \begin{aligned}
        \label{eq_soft_weights}
        w_{f} = 1 - \frac{e^{E_{f}(p)}}{e^{E_{b}(p)}+e^{E_{f}(p)}}\\
        w_{b} = 1 - \frac{e^{E_{b}(p)}}{e^{E_{b}(p)}+e^{E_{f}(p)}}\\
    \end{aligned}
\end{equation}
If both $E_{f}$ and $E_{b}$ are small, the area is more likely to be non-occluded in both direction.
Then the forward and backward errors should all be retained in the loss function, and after softmax, the corresponding $w_{b}$ and $w_{f}$ will be close to 0.5.
On the contrary, the larger the difference between the $E_{f}$ and $E_{b}$, the more likely this area is occluded in one direction.
Therefore, if $E_f$ is bigger than $E_b$, $w_b$ calculated by \Refeq{eq_soft_weights} will be smaller than $w_f$. 
Our weighted loss is then formulated as:

\begin{equation}
    \label{E2}
    \begin{aligned}
    L_{p1st}&= \sum_{p}\{w_{b}(p)\delta(E_{b}(p))+w_{f}(p)\delta(E_{f}(p))\}\\
    L_{p2nd}&= \sum_{d}\sum_{p}\{w_{b}(p)\delta(\nabla ^{d}I_{t}(p)-\nabla ^{d}I_{t-1}(F_{b}(p)+p))\\
            &+w_{f}(p)\delta(\nabla ^{d}I_{t}(p)-\nabla ^{d}I_{t+1}(F_{f}(p)+p))\}
    \end{aligned}
\end{equation}

\section{Experiments}

In this section, we validate the effectiveness of our method in public benchmark datasets of KITTI2012 and KITTI2015.
And we compare our method with existing deep learning-based methods.
For evaluation, we use end-point error (epe) which is defined as the average Euclidean distance between estimated and ground-truth optical flows.

\subsection{Datasets for Training}
\subsubsection*{KITTI}

We use KITTI raw data recordings to train our network, including 40864 samples in 65 training scenes and 3822 samples in 13 validation scenes.
The original images are downsampled to 832x256 taking into account the GPU memory and batch size restriction.

\subsection{Training details}
\subsubsection*{Pre-processing and data augmentations}

We adopt three kinds of data augmented methods: random scale cropping, random horizontal flipping, and normalization.
We first randomly flip the input images horizontally with a probability of 0.5.
Then we upscale each image, and the magnification factor is sampled between 1 and 1.1 according to a uniform distribution.
We also crop the scaled image to keep the same size as before.
Finally, we perform normalization with the variance of 0.5, and the mean of 0.5.

\subsubsection*{Optimizer}
Our network is trained using Adam optimizer \cite{KingmaAdam} with $\beta _{1}$ = 0.9 and $\beta _{2}$ = 0.999.
The initial learning rate is set to be $1e-4$, and is set to be $1e-5$ after 200K iterations.
The batch size is set to be 16, and we train our network for 300K iterations.

\subsubsection*{Hyperparameters}
The corresponding hyperparameters $[\lambda_{p1st}, \lambda_{p2nd}, \lambda_{s2nd}]$ are set to be $[0.06, 8, 10]$.
And the parameter $\alpha$ which controls the weight of edges in smooth loss is set to be 10, same as \cite{Wang}.
Moreover, we calculated the loss of the optical flow at different scales.
And different scales have different weights: $\lambda_{level1}=1.0, \lambda_{level_{i+1}}=\lambda_{level_{i}}/2\sqrt{2}$, which is similar to \cite{Godard2018}.
The hyperparameters $\epsilon$ and $\kappa$ in Charbonnier function are set to be 0.0001 and 2.
And the batch size is set to be 16.


\subsection{Results}
\begin{table}[htbp]
    \setlength{\abovecaptionskip}{-5pt}
    \caption{Quantitative evaluation of our method on different benchmarks.}
    \label{table-compare-with-others}
    \begin{center}
    \begin{tabular}{ccccccc}
    \toprule
    \specialrule{0em}{3pt}{3pt}
    \multirow{2}{*}{\makecell[c]{Method}} & \multicolumn{3}{c}{KITTI 2012} & \multicolumn{3}{c}{KITTI 2015} \\ \cmidrule(lr){2-4} \cmidrule(lr){5-7} & ALL & NOC & OCC & ALL & NOC & OCC \\
    \specialrule{0em}{3pt}{3pt}
    \midrule
    \specialrule{0em}{3pt}{3pt}
    FlowNetS+ft\cite{Fischer}  &  7.52 & - & - & - & - & -\\
    \specialrule{0em}{3pt}{3pt}
    FlowNet2\cite{Ilg}  &  4.09 & - & - & 10.06 & - & -\\
    \specialrule{0em}{3pt}{3pt}
    FlowNet2+ft\cite{Ilg}  & (1.28) & - & - & (2.3) & - & -\\
    \specialrule{0em}{3pt}{3pt}
    SpyNet+ft\cite{RanjanOptical}  &  8.25 & - & - & - & - & -\\
    \specialrule{0em}{3pt}{3pt}
    PWC-Net+ft\cite{SunPWC}  & (\textbf{1.45}) & - & - & (\textbf{2.16}) & - & -\\ 
    \hline
    \specialrule{0em}{3pt}{3pt}
    Back2Basic\cite{Yu2016}  & 11.3 & 4.3 &  & - & - & - \\
    \specialrule{0em}{3pt}{3pt}
    DSTFlow \cite{Ren2017} & - & 3.29 & 10.43 & - & 6.69 & 16.79 \\
    \specialrule{0em}{3pt}{3pt}
    UnFlow \cite{Meister2017} & 3.78 & 1.58 & - & 8.80 & 4.29 & - \\
    \specialrule{0em}{3pt}{3pt}
    OAFlow \cite{Wang} & \textbf{3.55} & - & - & 8.88 & - & - \\
    \specialrule{0em}{3pt}{3pt}
    Geonet \cite{Yin2018} & - & - & - & 10.81 & - & - \\
    \specialrule{0em}{3pt}{3pt}
    CC \cite{Ranjan2018} & - & - & - & 7.76 & - & - \\
    \specialrule{0em}{3pt}{3pt}
    Ours & 4.23 & \textbf{1.37} & \textbf{6.64} & \textbf{7.65} & \textbf{3.52} & \textbf{10.92} \\
    \bottomrule
    \end{tabular}
    \end{center}
    \footnotesize{Missing entries (-) indicate that the results are not reported for the respective method. Bold fonts highlight the best results among supervised and unsupervised methods respectively.}
\end{table}

\begin{figure*}[tbp]
    \centering
    \subfigure{
    \includegraphics[width=0.9\textwidth]{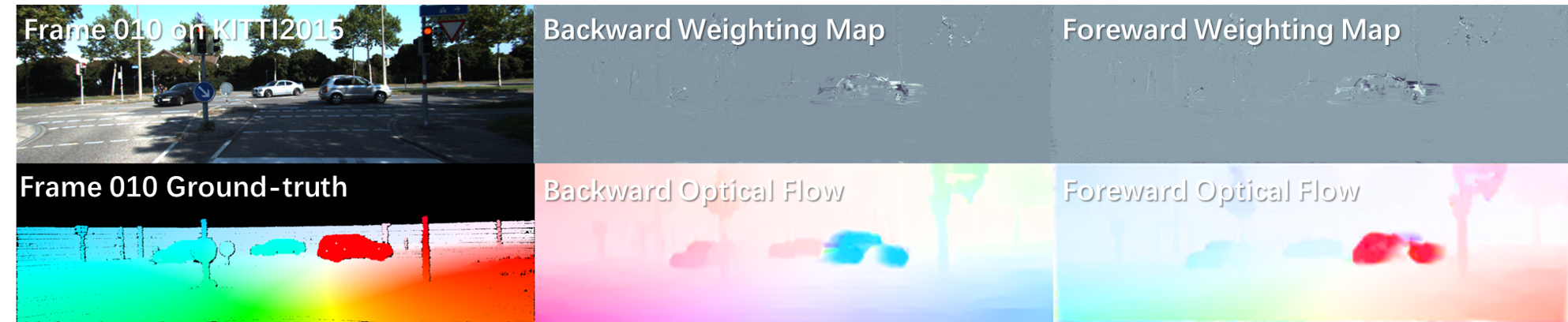}
    }
    \subfigure{
    \includegraphics[width=0.9\textwidth]{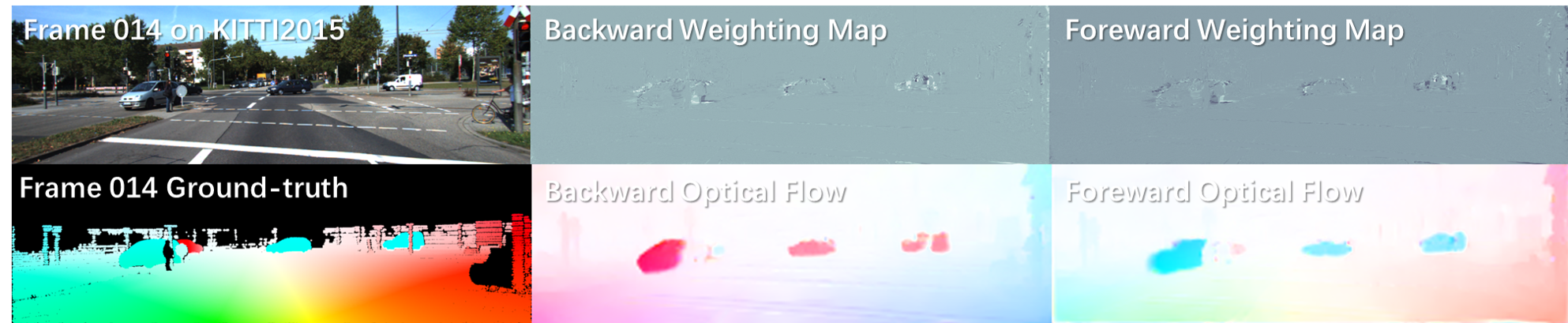}
    }
    \caption{Demonstration of the results on KITTI 2015 dataset. In the weighting map, a higher grayscale represents a higher weight.}
    \label{experiment-show}   
\end{figure*}

\begin{figure*}[tbp]
    \centering
    \includegraphics[width=0.9\textwidth]{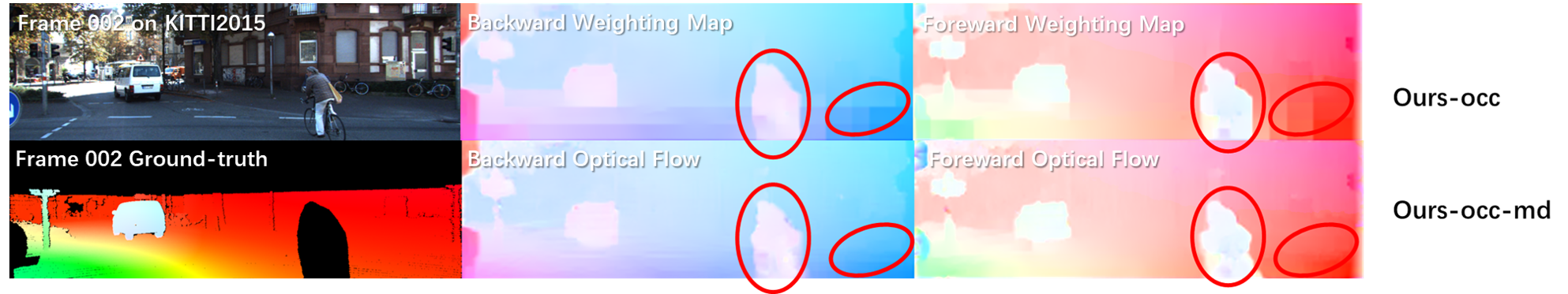}
    \caption{Demonstration of the effectiveness of multiple gradient directions. Ours-occ: only apply the collusion weighting; Ours-occ-md: apply occlusion weighting and multiple gradient directions.} 
    \label{experiment-md-show}   
\end{figure*}

The KITTI flow 2015 benchmark includes 200 training scenes and 200 test scenes, while the KITTI flow 2012 benchmark includes 194 training scenes and 195 test scenes.
Compared to KITTI flow 2012, KITTI flow 2015 includes dynamic scenes and the ground truth is established in a semi-automatic process \cite{Menze2015Object}.
We evaluate our approach on the training set for both datasets.
To better analyze the effect of the proposed method, the epe error is calculated in the occluded area and the non-occluded area respectively.

\Reftab{table-compare-with-others} shows that our approach has improved the performance in occluded areas significantly.
The epe is improved from 16.79 to 10.92 on KITTI 2015 and from 10.43 to 6.64 on KITTI 2012.
The proposed method outperforms the other methods in KITTI 2015, but slightly behind some work in KITTI 2012.
Compared with the baseline unsupervised method, our method is the best and even better than some of the existing supervised methods, \eg{FlowNetS+ft, SpyNet+ft}.
However, it is still behind the state-of-the-art supervised methods.

\Reffig{experiment-show} shows the qualitative results on KITTI2015.
The proposed method correctly captures the occluded area caused by the movement of vehicles.
It can be seen that the occluded area in forward weighting map is black while in the backward weighting map is white.

%
\Reffig{experiment-md-show} shows that calculating multiple gradient directions boosts the optical flow estimation on the silhouette of dynamic objects.

\begin{table}[htbp]
    \setlength{\abovecaptionskip}{-5pt}
    \caption{Ablation study of our method.}
    \label{table_ablation}
    \begin{center}
    \begin{tabular}{ccccccc}
    \toprule
    \specialrule{0em}{3pt}{3pt}
    \multirow{2}{*}{\makecell[c]{Method}} & \multicolumn{3}{c}{KITTI2012} & \multicolumn{3}{c}{KITTI2015} \\ \cmidrule(lr){2-4} \cmidrule(lr){5-7} & ALL & NOC & OCC & ALL & NOC & OCC \\
    \specialrule{0em}{3pt}{3pt}
    \midrule
    Ours-base & 4.62 & 1.91 & 6.91 & 8.17 & 4.20 & 11.31 \\
    \specialrule{0em}{3pt}{3pt}
    Ours-occ & 4.51 & 1.84 & 6.77 & 7.79 & 3.91 & \textbf{10.86} \\
    Ours-occ-md & \textbf{4.23} & \textbf{1.37} & \textbf{6.64} & \textbf{7.65} & \textbf{3.52} & 10.92 \\
    \bottomrule
    \end{tabular}
    \end{center}
\end{table}

\subsection{Ablation study}
In order to better reveal the performance of the proposed method, we conducted a series of ablation experiments, as shown in \Reftab{table_ablation}.
Our-base employs only the first and second-order photometric loss and smooth loss.
Our-occ employs the proposed weighting in the backward and forward photometric loss according to occlusion.
-md refers to the second-order photometric loss along multiple gradient directions.
The results show that after applying the proposed occlusion weighting, the epe of OCC and NOC both improve on KITTI 2015 and KITTI 2012.
Taking KITTI 2015 as an example, the epe of OCC improved from 11.31 to 10.86 and the epe of NOC improved from 4.20 to 3.91.
And after using multiple gradient directions, the epe of OCC improved from 6.77 to 6.64 and the epe of NOC improved from 1.84 to 1.37 on KITTI 2012.

\section{Conclusion}
In this paper, an occlusion aware optical flow estimation method based on unsupervised learning is proposed.
The proposed soft weighting method is able to handle the occlusion by using the prior knowledge that the occlusion is complementary in the forward and backward frame pairs.
By adding multiple gradient directions in the loss calculation, the optical flow at the edges is clearer and finer.
The experiments show the proposed approach outperforms the baseline methods on KITTI benchmark datasets.
Our method does not completely eliminate the occluded areas, but minimizes the impact of these occluded areas in a soft way.
In the future, the timing relationships between video frames have to be incorporated. 
And more geometric constraints between optical flow and depth and pose could be applied.

\bibliographystyle{IEEEtran}
\bibliography{IEEEabrv,root}

\end{document}